\documentclass{salesforce_ai_research}

\usepackage[T1]{fontenc}
\usepackage[utf8]{inputenc}
\usepackage{amsmath}
\usepackage{amsfonts}
\usepackage{amssymb}
\usepackage{nicefrac}
\usepackage{microtype}
\usepackage{graphicx}
\usepackage{booktabs}
\usepackage{multirow}
\usepackage{longtable}
\usepackage{array}
\usepackage{pifont}
\usepackage{verbatim}
\usepackage{enumitem}
\setlist{nosep,leftmargin=*,topsep=2pt,partopsep=0pt,parsep=0pt,itemsep=0pt}

\newcommand{\MOne}{M\textsubscript{1}}
\newcommand{\MTwo}{M\textsubscript{2}}
\newcommand{\MThree}{M\textsubscript{3}}
\newcommand{\MFour}{M\textsubscript{4}}
\newcommand{\MFive}{M\textsubscript{5}}
\newcommand{\MSix}{M\textsubscript{6}}
\newcommand{\ERS}{\textsc{EGI}}

\newcommand{\bench}[1]{\textsc{#1}}

\newcommand{\ebench}{\textsc{EGI-Bench}}
\newcommand{\ebenchcli}{\texttt{egi-bench}}

\title{\ebench{}: An Enterprise General Intelligence Benchmark Suite}

\author{
Zhiwei Liu\thanks{Equal contribution.},\quad Zhujun Lan\footnotemark[1],\quad Jielin Qiu,\quad Zixiang Chen,\quad Liangwei Yang,\\
\bfseries Ming Zhu,\quad Akshara Prabhakar,\quad Wenting Zhao,\quad John Emmons,\quad Jason Wu,\\
\bfseries Shafiq Rayhan Joty,\quad Junnan Li,\quad Shelby Heinecke,\quad Silvio Savarese,\quad Huan Wang\thanks{Corresponding author: \texttt{huan.wang@salesforce.com}.}\\[4pt]
\normalfont Salesforce AI Research
}

\begin{document}

\maketitle

\begin{abstract}
Enterprise adoption of LLMs in agentic workflows is gated by
evaluations that span a wide arc of abilities: CRM operations,
heterogeneous-document retrieval, personal data understanding,
multi-turn planning, tool use, function calling, long-horizon
research, and long-context software engineering. These evaluations
live today in separate repositories with divergent runners,
dependencies, and scoring conventions, which makes cross-benchmark
comparison difficult and obscures what a single aggregate number
actually represents. We release \ebench{}, a unified LLM agent benchmark
suite that consolidates eight enterprise-facing agent benchmarks
(\bench{CRMArena}, \bench{HERB}, \bench{PersonaBench},
\bench{UserBench}, \bench{MCP-Universe}, \bench{MCPEval},
\bench{Benchforce}, \bench{LoCoBench}) behind a single command-line
interface with two runtime-budgeted execution modes. On top of the
suite we define the \emph{Enterprise General Intelligence} (\ERS{})
Score, a six-dimensional framework
(\MOne--\MSix{}) whose buckets are populated by \emph{fine-grained
sub-metrics} routed from each benchmark rather than by benchmark-level
averages. We describe the suite, the sub-metric routing, and the
quick/deep budgets, and report results for fourteen frontier and
cost-efficient LLMs spanning the GPT, Claude, and Gemini families
under quick mode. Top-line \ERS{} ranges from $39.6$ (GPT-4o~mini) to
$65.6$ (\texttt{gpt-5.5}), and no single model dominates every
capability axis: \texttt{gpt-5.5} leads task outcome, planning, and
retrieval grounding; Claude Sonnet~4.5 leads tool actioning and
safety/abstention; and Claude Haiku~4.5 leads engineering quality. Code is available at
\url{https://github.com/SalesforceAIResearch/egi-bench}.

\end{abstract}

\section{Introduction}
\label{sec:intro}

Large language models are now deployed inside agentic systems that
plan, use tools, and act inside real software
environments~\citep{yao2022react,schick2023toolformer,qin2023toolllm}.
In the enterprise, these systems answer customer-support tickets,
search internal documents, navigate CRM objects, plan travel, invoke
business APIs, and maintain large codebases. For any given
deployment, the question a release-gating team asks is whether the
underlying LLM is \emph{ready} for these workflows: can it produce
correct business outcomes, call tools reliably, plan efficiently,
ground answers in retrieved evidence, abstain when appropriate, and
produce artifacts of acceptable engineering quality? Answering that
question requires many signals at once, and no single benchmark
captures all of them. \ebench{} measures the LLM that drives these
workflows; each contributing benchmark exercises it through its own
native scaffold (ReAct loops, multi-turn dialog, function-calling
harnesses), and the resulting \ERS{} score is therefore a model-level
signal across enterprise-shaped tasks.

Despite this demand, existing benchmarks each cover a narrow slice of
the space. The agent-evaluation literature has grown rapidly but is
fragmented along capability lines.
\bench{AgentBench}~\citep{liu2023agentbench} and
\bench{AgentBoard}~\citep{ma2024agentboard} evaluate general agentic
behavior across heterogeneous environments; \bench{GAIA}~\citep{mialon2023gaia}
targets general AI-assistant questions with verifiable answers.
\bench{SWE-bench}~\citep{jimenez2023swebench} focuses on resolving
real-world GitHub issues; \bench{WebArena}~\citep{zhou2023webarena}
places agents inside a realistic web environment.
\bench{ToolLLM}~\citep{qin2023toolllm}, the Berkeley Function Calling
Leaderboard~\citep{bfcl2024}, and \bench{MINT}~\citep{wang2023mint}
probe tool use and multi-turn tool-augmented interaction;
$\tau$-bench~\citep{yao2024taubench} measures tool-agent-user dialog in
realistic domains. Holistic LLM evaluations such as
\bench{HELM}~\citep{liang2022helm} aggregate scores across many tasks
but are not agent-centric. In the enterprise setting specifically,
recent benchmarks cover CRM workflows~\citep{crmarena2024,crmarenapro2026}, deep search
over heterogeneous enterprise data~\citep{herb2025}, personalized QA
over private user data~\citep{personabench2025}, multi-turn
user-facing dialog~\citep{userbench2025}, function
calling~\citep{benchforce2025}, MCP-based tool
use~\citep{mcpuniverse2025,mcpeval2025}, and long-context software
engineering~\citep{locobench2025}. Each of these benchmarks lives in
its own repository, with its own runner, dependencies, and scoring
conventions.

Because these benchmarks live in separate repositories with divergent
runners and scoring conventions, a practitioner who wants a single
\ERS{} signal today must run many tools, reconcile
their outputs by hand, and decide how to weigh them. Naively averaging
each benchmark's reported overall score dilutes signal: a benchmark that
internally averages nine sub-scores contributes one number, while a
benchmark with a single sub-metric contributes the same one number.
Worse, each benchmark's ``overall'' collapses sub-metrics that probe
very different capabilities (e.g.\ policy-violation detection vs.\
named-entity disambiguation inside
\bench{CRMArena}~\citep{crmarena2024}). The result is that existing
leaderboards cannot answer capability-grounded questions like ``is this
model safe enough to ship?'' or ``does it plan multi-step workflows
reliably?''. Table~\ref{tab:benchmark-comparison} makes this gap
concrete: the four most prominent agent-benchmark reference points
each cover only a strict subset of the six \ERS{}
dimensions we argue are needed for release decisions. This is the gap
\ebench{} fills.

\paragraph{Contributions.} This report describes \ebench{}, a unified
LLM agent benchmark suite and scoring framework for enterprise-shaped
tasks. Our contributions are:

\begin{itemize}
  \item \textbf{Unified suite.} We consolidate eight enterprise-facing
  LLM agent benchmarks~\citep{crmarena2024,herb2025,personabench2025,
  userbench2025,mcpuniverse2025,mcpeval2025,benchforce2025,
  locobench2025} into one repository behind the
  \ebenchcli{} CLI, with per-benchmark virtual environments, a shared
  LiteLLM model-routing proxy, and a single
  \texttt{run --model \dots\ --mode \{quick,deep\}} entry point. Each
  benchmark retains its own scaffold (ReAct, MCP client, multi-turn
  dialog, etc.); the unit of evaluation is the LLM.
  \item \textbf{Fine-grained sub-metric routing.} Rather than averaging
  benchmark-level scores, we route \emph{individual sub-metrics} into a
  six-dimensional \ERS{} framework (\MOne--\MSix{}) that
  captures task-outcome accuracy, tool actioning, planning, retrieval
  grounding, safety/abstention, and engineering quality. The routing is
  declarative and auditable (Appendix~\ref{sec:routing}).
  \item \textbf{Runtime-budgeted modes.} Quick mode targets
  ${\leq}180$~min total wall-clock; deep mode targets ${\sim}12$~h.
  Which categories or tasks get subsetted is determined by a
  per-benchmark runtime budget rather than by hand-picked coverage.
  \item \textbf{Open release with a fourteen-model benchmark.} We
  release the suite, configuration, and full quick-mode results across
  fourteen frontier and cost-efficient LLMs spanning the GPT, Claude,
  and Gemini model families. The results expose capability-level
  specialization that is invisible in any single benchmark's overall
  score.
\end{itemize}

The rest of the report surveys related evaluation work
(Section~\ref{sec:related}), positions \ebench{} against comparable
benchmarks (Section~\ref{sec:comparison}), defines the \MOne--\MSix{}
framework and the eight benchmark contributors
(Section~\ref{sec:framework}), reports results and analysis
(Sections~\ref{sec:results}--\ref{sec:analysis}), and lists limitations
(Section~\ref{sec:limitations}). The sub-metric routing table and
execution-mode details are deferred to
Appendices~\ref{sec:routing} and~\ref{sec:modes}.

\section{Related Work}
\label{sec:related}

We situate \ebench{} relative to four lines of work: holistic LLM
evaluations, general agent benchmarks, tool-use and function-calling
benchmarks, and enterprise-facing agent benchmarks.

\paragraph{Holistic LLM evaluations.} \bench{HELM}~\citep{liang2022helm}
proposed a multi-metric view of language-model evaluation, separating
accuracy, robustness, calibration, and other axes across many tasks.
While HELM is primarily task-level rather than agent-level, it established the principle that a single benchmark should expose \emph{multiple interpretable dimensions}. \ebench{} adopts the same
spirit but applies it to LLM evaluation in enterprise agentic
workflows by defining six enterprise-relevant dimensions
(\MOne--\MSix{}) and routing each contributing benchmark's native
sub-metrics into them, rather than averaging benchmark-level scores.

\paragraph{General agent benchmarks.}
\bench{AgentBench}~\citep{liu2023agentbench} evaluates LLM agents across
eight heterogeneous environments (code, games, web, reasoning, etc.)
and reports per-environment success rate with an overall average; it
does not expose a cross-environment capability decomposition, so
planning, tool use, and abstention collapse into a single scalar per
environment. \bench{AgentBoard}~\citep{ma2024agentboard} introduces
\emph{progress rate}, a fine-grained mid-trajectory metric across nine
environments --- the closest prior art to our ``don't just report the
final number'' philosophy, but operating within a single benchmark
family rather than composing across independently authored benchmarks.
\bench{GAIA}~\citep{mialon2023gaia} benchmarks general AI-assistant
questions with verifiable ground truth and public/private test sets;
the only axis of decomposition is its three-level difficulty partition
(Level 1--3), and a failed web-search step and a failed arithmetic step
are indistinguishable in the scored output.
\bench{WebArena}~\citep{zhou2023webarena} places agents in a realistic,
self-hosted web environment with full interaction traces and reports
end-to-end task success. These benchmarks target breadth of capability,
not enterprise readiness: none of them decomposes scores into
enterprise-interpretable buckets such as safety/abstention or
engineering quality, and none composes signals across independently
authored benchmark families.

\paragraph{Tool-use and function-calling benchmarks.}
\bench{ToolLLM}~\citep{qin2023toolllm} introduced ToolBench with
16{,}000+ real-world APIs and trained a tool-augmented LLM; its
headline metric is a pass-rate-with-retries aggregated across APIs.
The Berkeley Function Calling Leaderboard (BFCL)~\citep{bfcl2024}
standardizes evaluation of function calling across API families and
provides the most fine-grained tool-use report to date: AST vs.\
executable evaluation, simple / multiple / parallel / parallel-multiple
settings, multi-turn and multi-step categories, language axes (Python,
REST, SQL, Java, JavaScript), Function Relevance Detection (an explicit
abstention test in which the correct response to an irrelevant toolset
is to call nothing), and cost-per-1000-calls with latency as
first-class signals. BFCL's taxonomy on the tool-calling axis is
stronger than what \ebench{}'s MCP contributors currently surface;
\ebench{}'s advantage is the enterprise context that surrounds the
tool calls, not the tool-calling depth itself.
\bench{MINT}~\citep{wang2023mint} measures multi-turn interaction with
tools under language feedback, reporting success at $k$ tool calls and
tool-type breakdowns. $\tau$-bench~\citep{yao2024taubench} and its
dual-control follow-on $\tau^2$-bench~\citep{barres2025tau2bench}
target tool-agent-user interaction in realistic retail, airline, and
(in $\tau^2$) telecom domains, with pass$^k$ ($k$ up to $8$) as the
headline metric plus an auto error-identification pipeline that
classifies failures by fault assignment (agent / user / environment)
and fault type (\texttt{used\_wrong\_tool},
\texttt{used\_wrong\_tool\_argument}, \texttt{took\_unintended\_action},
\texttt{goal\_partially\_completed}). 
This multi-trial reliability
signal is missing from \ebench{} v1 and is concrete future work. Within
\ebench{}, tool-use signals come from
\bench{MCP-Universe}~\citep{mcpuniverse2025} and
\bench{MCPEval}~\citep{mcpeval2025}, which instantiate the Model
Context Protocol~\citep{mcpspec2024} server/agent contract, while
\bench{Benchforce}~\citep{benchforce2025} complements these with
WebSocket-driven enterprise function-calling environments. Where prior
tool-use benchmarks expose a single success rate, \ebench{} routes the
component metrics (tool-name match, parameter match, ordering, strict
vs.\ flexible success) into separate enterprise buckets so that drift,
over-action, and planning-order errors are surfaced independently.

\paragraph{Code and long-context benchmarks.}
\bench{SWE-bench}~\citep{jimenez2023swebench} evaluates whether LLMs
can resolve real-world GitHub issues in 12 Python repositories using
unit tests as verifiers; its per-instance pipeline surfaces four
signals (applied / tests-pass / regression / resolved), with the
headline leaderboard reporting the resolve rate. Hidden-test ground
truth certified by the original issue authors makes SWE-bench the
stronger instrument for pure coding correctness, but it does not
measure code quality, patch-size minimization, planning or search
behavior separate from the final patch, or tool-call efficiency.
\bench{LoCoBench}~\citep{locobench2025} extends software-engineering
evaluation to long-context scenarios spanning ten programming languages
and context lengths up to ${\sim}1$M tokens, using a four-dimensional
rubric (functional correctness, software engineering excellence, code
quality, long-context utilization). \ebench{} uses \bench{LoCoBench} as
its software-engineering contributor and routes its four dimensions
into \MOne{}, \MThree{}, and \MSix{}; the rubric is softer than
SWE-bench's hidden tests, which we acknowledge as a trade-off.

\paragraph{Enterprise-facing agent benchmarks.} A newer wave of
benchmarks targets enterprise workflows specifically.
\bench{CRMArena}~\citep{crmarena2024,crmarenapro2026} evaluates Salesforce-style CRM
operations (case routing, analytical queries, policy-violation
detection) against a sandbox org. \bench{HERB}~\citep{herb2025}
benchmarks multi-hop retrieval over realistic, heterogeneous enterprise
corpora (documents, Slack, meeting transcripts, URLs) and mixes
answerable and unanswerable queries to test anti-hallucination.
\bench{PersonaBench}~\citep{personabench2025} measures personalized QA
over synthetic private user data with controllable noise injection.
\bench{UserBench}~\citep{userbench2025} exercises multi-turn travel-planning
dialog with user-preference satisfaction as the reward signal.
Each of these benchmarks is valuable on its own;
what is missing is a principled way to read them \emph{together} as an
\ERS{} signal.

\paragraph{Cross-domain aggregation.} Prior practice aggregates
cross-benchmark signal either by picking a headline benchmark per
capability or by averaging benchmark-level overall scores. Both lose
information. \ebench{} instead routes individual sub-metrics into the
\MOne--\MSix{} framework (Appendix~\ref{sec:routing}), preserves per-benchmark
sub-scores for diagnosis (Section~\ref{sec:results}), and optionally
applies release-gate floors on \MFive{} (safety/abstention) and
\MTwo{} (tool actioning) so that a single aggregate number cannot hide
a capability regression that matters for deployment.

\section{Benchmark Comparison and Positioning}
\label{sec:comparison}

We position \ebench{} against four representative open-source
agent/LLM benchmarks: $\tau^2$-bench~\citep{barres2025tau2bench} (the
dual-control follow-on to the original
$\tau$-bench~\citep{yao2024taubench}),
SWE-bench~\citep{jimenez2023swebench}, GAIA~\citep{mialon2023gaia}, and
the Berkeley Function Calling Leaderboard (BFCL)~\citep{bfcl2024}. The
head-to-head coverage matrix is shown in
Table~\ref{tab:benchmark-comparison}; here
we discuss the trade-offs the matrix encodes.

\begin{table*}[t]
\centering
\footnotesize
\setlength{\tabcolsep}{4pt}
\renewcommand{\arraystretch}{1.05}
\caption{Domain coverage, task count, and \MOne--\MSix{} enterprise-metric
coverage across representative agent/LLM benchmarks. \ding{51}{} = the
benchmark evaluates this dimension as a core part of its design;
$\sim${} = exercised as part of tasks but not a primary axis;
\ding{55}{} = outside scope.}
\label{tab:benchmark-comparison}
\begin{tabular}{@{}
  >{\raggedright\arraybackslash}p{2.9cm}
  >{\raggedright\arraybackslash}p{3.1cm}
  >{\raggedright\arraybackslash}p{2.6cm}
  >{\raggedright\arraybackslash}p{2.2cm}
  >{\raggedright\arraybackslash}p{2.0cm}
  >{\raggedright\arraybackslash}p{2.4cm}
@{}}
\toprule
 & \ebench{} & $\tau^2$-bench & SWE-bench & GAIA & BFCL \\
\midrule
Domains                        & 8 benchmarks; $\sim$20 sub-domains         & 3 domains (dual-control): retail, airline, telecom & 1 domain: SWE (Python) & 1 domain: generalist & 1 domain: function calling \\
Tasks                          & $\sim$1{,}500+                             & a few hundred / domain             & 2{,}294 / 500 / 300         & $\sim$466                       & $\sim$2{,}000 \\
\midrule
\MOne{} Task Outcome           & \multicolumn{1}{c}{\ding{51}}              & \multicolumn{1}{c}{\ding{51}}      & \multicolumn{1}{c}{\ding{51}} & \multicolumn{1}{c}{\ding{51}} & \multicolumn{1}{c}{\ding{51}} \\
\MTwo{} Tool Actioning         & \multicolumn{1}{c}{\ding{51}}              & \multicolumn{1}{c}{\ding{51}}      & \multicolumn{1}{c}{$\sim$}  & \multicolumn{1}{c}{$\sim$}    & \multicolumn{1}{c}{\ding{51}} \\
\MThree{} Planning             & \multicolumn{1}{c}{\ding{51}}              & \multicolumn{1}{c}{\ding{51}}      & \multicolumn{1}{c}{$\sim$}  & \multicolumn{1}{c}{\ding{51}} & \multicolumn{1}{c}{\ding{51}} \\
\MFour{} Retr./Grounding       & \multicolumn{1}{c}{\ding{51}}              & \multicolumn{1}{c}{$\sim$}         & \multicolumn{1}{c}{$\sim$}  & \multicolumn{1}{c}{\ding{51}} & \multicolumn{1}{c}{$\sim$}    \\
\MFive{} Safety/Abstention     & \multicolumn{1}{c}{\ding{51}}              & \multicolumn{1}{c}{\ding{51}}      & \multicolumn{1}{c}{\ding{55}} & \multicolumn{1}{c}{\ding{55}} & \multicolumn{1}{c}{\ding{51}} \\
\MSix{} Engineering            & \multicolumn{1}{c}{\ding{51}}              & \multicolumn{1}{c}{\ding{55}}      & \multicolumn{1}{c}{\ding{51}} & \multicolumn{1}{c}{\ding{55}} & \multicolumn{1}{c}{\ding{55}} \\
\midrule
\textbf{\# M-dimensions (\ding{51})}  & \multicolumn{1}{c}{\textbf{6}}            & \multicolumn{1}{c}{4}              & \multicolumn{1}{c}{2}       & \multicolumn{1}{c}{3}         & \multicolumn{1}{c}{4} \\
\bottomrule
\end{tabular}
\end{table*}

Each of the four comparison benchmarks is a strong instrument on a
narrow subset of axes: $\tau^2$-bench on dialog reliability and
pass$^k$-style multi-trial correctness with a dual-control user
simulator, SWE-bench on code-resolution correctness with hidden-test
ground truth, GAIA on multi-step generalist-assistant tasks, and BFCL
on tool-calling taxonomy and cost/latency telemetry. \ebench{}'s
contribution is not to beat any of them on its home axis but to
\emph{cover all six} \ERS{} dimensions through a
composition of eight independently authored benchmarks, route their
sub-metrics into a shared frame (Appendix~\ref{sec:routing}), and add
release-gate semantics via configurable \MFive{} and \MTwo{} floors.
Practitioners who need the maximum-rigor instrument on a single axis
should continue to use the specialist benchmark; practitioners who
need a capability-grounded readiness score for an enterprise deployment
should use \ebench{}.
\section{The \ERS{} Framework}
\label{sec:framework}

We evaluate LLMs along six enterprise-level metrics and combine them
into a single \ERS{} score. The metrics,
definitions, and weights are summarized in Table~\ref{tab:m-definitions}.

\begin{table}[!ht]
\centering
\small
\setlength{\tabcolsep}{5pt}
\renewcommand{\arraystretch}{1.15}
\caption{The six enterprise metrics (M1--M6) and their weights in the
\ERS{} score. Weights are inherited from the internal PR\,\#28
proposal and are applied uniformly across the suite.}
\label{tab:m-definitions}
\begin{tabular}{@{}l c >{\raggedright\arraybackslash}p{10.5cm}@{}}
\toprule
Metric & Weight & What it measures \\
\midrule
\MOne{} Task Outcome Accuracy             & 28\% & Did the agent produce the correct business outcome or final answer? \\
\MTwo{} Tool Actioning Reliability        & 22\% & Correct tool selection, parameters, ordering; no extraneous calls. \\
\MThree{} Planning \& Workflow            & 16\% & Multi-step and multi-turn execution that is step-efficient. \\
\MFour{} Retrieval \& Evidence Grounding  & 14\% & Grounded answers, correct citations, effective context use. \\
\MFive{} Safety, Abstention \& Policy     & 14\% & Abstain when appropriate, flag policy violations, no unsafe actions. \\
\MSix{} Engineering \& Maintainability    &  6\% & Code quality and structural/process discipline. \\
\midrule
\multicolumn{3}{@{}l@{}}{$\ERS{} = 0.28\,\MOne{} + 0.22\,\MTwo{} + 0.16\,\MThree{} + 0.14\,\MFour{} + 0.14\,\MFive{} + 0.06\,\MSix{}$} \\
\bottomrule
\end{tabular}
\end{table}

These six metrics capture capabilities that enterprise stakeholders
evaluate independently: \MOne{} asks ``did the agent produce the correct
answer?''; \MTwo{} asks ``did it use tools correctly?''; \MThree{}
asks ``did it plan efficiently across steps and turns?''; \MFour{}
asks ``are its answers grounded in retrieved evidence?''; \MFive{}
asks ``does it abstain and obey policy when it should?''; and \MSix{}
asks ``is the artifact it produced of acceptable engineering
quality?''. The weights $0.28/0.22/0.16/0.14/0.14/0.06$ bias the score
toward the outcomes that matter most in production deployments (task
accuracy and tool reliability) while keeping a non-trivial safety
floor. Each metric is computed over the outputs the LLM produces
inside the contributing benchmark's native scaffold.

To compute \ERS{}, each sub-metric is first normalized to $[0, 100]$;
lower-is-better sub-metrics (e.g.\ step counts in \bench{MCP-Universe})
are clipped against a stable ceiling and inverted. For each bucket
$M_k$, its score is the (equal-weighted by default) mean of the
normalized sub-metrics routed into it. The aggregate is then
\begin{equation*}
\begin{split}
\ERS{} ={}& 0.28\,\MOne{} + 0.22\,\MTwo{} + 0.16\,\MThree{} \\
         &{} + 0.14\,\MFour{} + 0.14\,\MFive{} + 0.06\,\MSix{}.
\end{split}
\end{equation*}
When a sub-metric is missing (e.g.\ a benchmark run failed), we reweight
the remaining sub-metrics \emph{within} that bucket only; we never
globally reweight across \MOne--\MSix{}. If an entire benchmark is
missing, all sub-metrics it contributes are skipped and affected
buckets are renormalized locally. This keeps buckets comparable while
signalling missing coverage honestly.

Beyond the aggregate, \MFive{} (safety/abstention) and \MTwo{} (tool
actioning) can optionally gate release: if either drops below a
configurable threshold (default $70$), the run is flagged \emph{not
release-ready} regardless of \ERS{}. Release-gate floors are reported
alongside \ERS{} but do not modify the score itself. For the routing to
be well-posed, every benchmark must contribute to at least one bucket
and every bucket must be populated by at least two routed sub-metrics.
\MSix{} draws both of its sub-metrics (\texttt{software\_engineering},
\texttt{code\_quality}) from a single benchmark (\bench{LoCoBench});
all other buckets receive sub-metrics from three or more benchmarks.
The full coverage view
is deferred to Appendix~\ref{sec:app-bench-tables},
Table~\ref{tab:coverage}, and the sub-metric routing is described in
Appendix~\ref{sec:routing}.

The eight benchmarks that populate \MOne--\MSix{} are:
\bench{CRMArena}~\citep{crmarena2024,crmarenapro2026}
for Salesforce CRM operations (nine task types, including policy
violation and knowledge QA); \bench{HERB}~\citep{herb2025} for deep
search over heterogeneous enterprise data with answerable/unanswerable
splits; \bench{PersonaBench}~\citep{personabench2025} for personalized
QA over synthetic private user data under controllable noise;
\bench{UserBench}~\citep{userbench2025} for multi-turn travel-planning
dialog with a 20-turn budget;
\bench{MCP-Universe}~\citep{mcpuniverse2025} for Model Context
Protocol~\citep{mcpspec2024} tool use across six server ecosystems
under a ReAct agent; \bench{MCPEval}~\citep{mcpeval2025} for
auto-synthesized tool-call tasks with strict/flexible matching;
\bench{Benchforce}~\citep{benchforce2025} for enterprise function
calling (hospital, HR, order-management); and \bench{LoCoBench}~\citep{locobench2025} for
long-context software engineering across ten languages at context
lengths up to 1\,M tokens. Appendix~\ref{sec:app-bench-tables} reports
per-benchmark task counts under quick and deep modes.
\section{Results}
\label{sec:results}

\begin{figure*}[!t]
\centering
{\centering
\footnotesize
\setlength{\tabcolsep}{5pt}
\renewcommand{\arraystretch}{1.10}
\captionof{table}{Main results: \ERS{} score and its
\MOne--\MSix{} constituents across fourteen frontier and cost-efficient
LLMs under quick mode. Rows sorted by \ERS{}. Scores are
0--100; higher is better. Bold = column best.}
\label{tab:main-results}
\begin{tabular}{@{}l c c c c c c c c@{}}
\toprule
Model & Mode & \ERS & \MOne & \MTwo & \MThree & \MFour & \MFive & \MSix \\
\midrule
\texttt{gpt-5.5}      & quick & \textbf{65.61} & \textbf{82.95} & 79.56 & \textbf{30.89} & \textbf{54.86} & 63.47 & 56.23 \\
Claude Sonnet 4.5     & quick & 65.25 & 69.89 & \textbf{87.17} & 27.48 & 53.04 & \textbf{80.28} & 57.34 \\
Gemini 3.1 Pro Prev.  & quick & 63.51 & 78.45 & 76.82 & 19.57 & 54.43 & 75.56 & 55.30 \\
Claude Opus 4.5       & quick & 63.31 & 74.29 & 75.58 & 27.95 & 52.08 & 75.56 & 58.98 \\
GPT-4o                & quick & 59.38 & 66.18 & 84.22 & 23.88 & 39.47 & 53.61 & 51.95 \\
Claude Haiku 4.5      & quick & 59.13 & 69.02 & 73.98 & 24.05 & 48.61 & 65.69 & \textbf{61.28} \\
Gemini 2.5 Pro        & quick & 59.10 & 76.05 & 65.60 & 27.16 & 50.58 & 60.83 & 57.25 \\
GPT-4.1               & quick & 58.31 & 69.88 & 81.60 & 27.36 & 44.99 & 49.03 & 54.19 \\
GPT-5 mini            & quick & 51.91 & 59.10 & 67.74 & 21.75 & 43.05 & 54.17 & 56.12 \\
Gemini 2.5 Flash      & quick & 51.36 & 59.25 & 64.10 & 24.72 & 40.99 & 54.58 & 55.48 \\
Claude 3.5 Haiku      & quick & 50.78 & 48.16 & 75.81 & 18.39 & 44.35 & 58.61 & 54.36 \\
Gemini 2.0 Flash      & quick & 49.29 & 48.57 & 72.53 & 20.65 & 42.54 & 51.67 & 53.96 \\
GPT-5 nano            & quick & 45.14 & 42.38 & 72.05 & 15.68 & 43.31 & 41.25 & 51.25 \\
GPT-4o mini           & quick & 39.56 & 30.91 & 64.45 & 16.66 & 39.33 & 39.17 & 51.14 \\
\bottomrule
\end{tabular}
\par}

\vspace{1.5em}
\includegraphics[width=\textwidth]{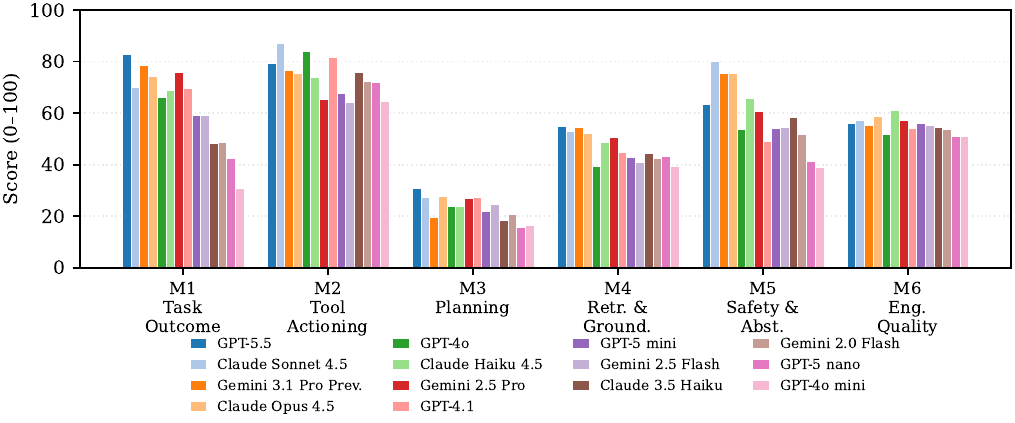}
\caption{\MOne{}--\MSix{} scores across fourteen LLMs under quick
mode. Models are ordered by \ERS{}; each bucket is a cluster of bars.
The \MThree{} (Planning) dimension is uniformly low across models, while
\MTwo{} (Tool Actioning) is the strongest axis for nearly every frontier
model. The best \MTwo{} (Tool Actioning, 87.17) and \MFive{} (Safety \&
Abstention, 80.28) are held by Claude Sonnet~4.5, while the best
\MOne{} (Task Outcome, 82.95), \MThree{} (Planning, 30.89), and
\MFour{} (Retrieval \& Grounding, 54.86) are all held by
\texttt{gpt-5.5}, illustrating that top-line \ERS{} can mask
capability-level specialization.}
\label{fig:m16-bars}
\end{figure*}

Table~\ref{tab:main-results} reports \ERS{} and its \MOne--\MSix{}
constituents for fourteen frontier and cost-efficient LLMs under
quick mode. Figure~\ref{fig:m16-bars} compares all fourteen models on
each \MOne--\MSix{} dimension as grouped bars, and
Figure~\ref{fig:m16-heatmap} renders the same matrix as a heatmap so
per-model specialization patterns are visible at a glance.

Across the fourteen models, \texttt{gpt-5.5} leads on overall \ERS{}
($65.61$), driven by best-in-class \MOne{} (Task Outcome, $82.95$),
\MThree{} (Planning, $30.89$), and \MFour{} (Retrieval \& Grounding,
$54.86$). Claude Sonnet~4.5 sits a hair behind at \ERS{} $65.25$ and
tops \MTwo{} (Tool Actioning, $87.17$) and \MFive{} (Safety \&
Abstention, $80.28$) -- the only safety score above $76$ in the
roster. Gemini~3.1 Pro Preview and Claude Opus~4.5 round out the
top tier at \ERS{} $63.51$ and $63.31$, with Gemini~3.1's strong
\MOne{} of $78.45$ second only to \texttt{gpt-5.5}. The most striking
specialization signal sits at rank~6: Claude Haiku~4.5's \MSix{}
(Engineering Quality, $61.28$) is column-best in the entire roster
despite its mid-pack \ERS{}. The cost-efficient tier (GPT-4o mini,
GPT-5 nano, Gemini 2.0/2.5 Flash) trails the frontier tier on \ERS{}
by roughly $6$--$26$ points, with the GPT-5 mini / Gemini~2.5 Flash
boundary a near-tie at $51.91$ vs $51.36$. \MThree{} is uniformly the
weakest dimension ($16$--$31$ across all fourteen models), driven by
\bench{UserBench}'s low per-environment reward, indicating multi-turn
planning remains an open frontier.
The heatmap in Figure~\ref{fig:m16-heatmap} makes these
per-row/per-column patterns directly visible: hot cells flag weak
scores and cool cells mark strong ones, so \MThree{} forms a uniformly
hot column and \MTwo{} a uniformly cool one, while the coolest cell in
each row marks the model's strongest dimension.

\section{Analysis}
\label{sec:analysis}

\begin{figure*}[!t]
\centering
\begin{minipage}[t]{0.48\textwidth}
\centering
\includegraphics[width=\linewidth]{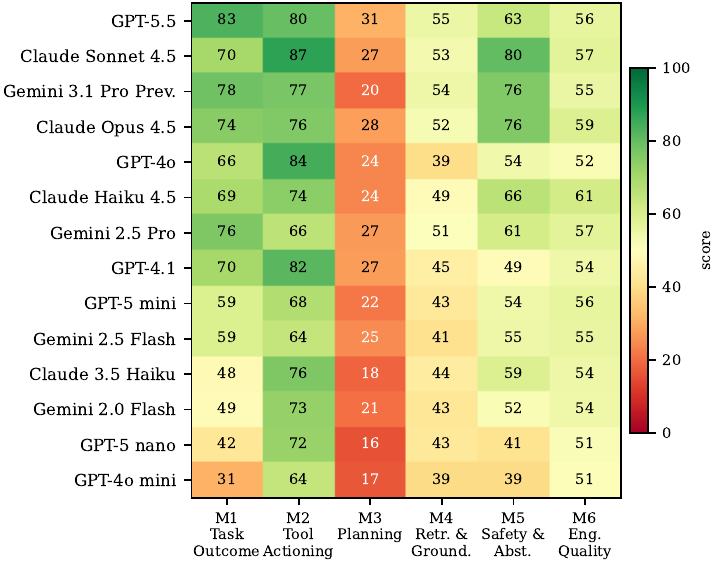}
\caption{\MOne{}--\MSix{} score matrix (14 models $\times$ 6 metrics),
sorted by \ERS{}. Hot cells flag weak scores, while cool cells mark
strong performance. \MThree{} (Planning) is therefore the hottest
column, since no model surpasses $31$; \MTwo{} (Tool Actioning) is the
coolest, with most frontier models performing above $70$. Within each
row, the coolest cell identifies that model's strongest dimension
(\emph{e.g.}\ \texttt{gpt-5.5} is coolest on \MOne{}, Claude
Sonnet~4.5 on \MTwo{} and \MFive{}, and Claude Haiku~4.5 holds the
column-best on \MSix{}).}
\label{fig:m16-heatmap}
\end{minipage}\hfill
\begin{minipage}[t]{0.48\textwidth}
\centering
\includegraphics[width=\linewidth]{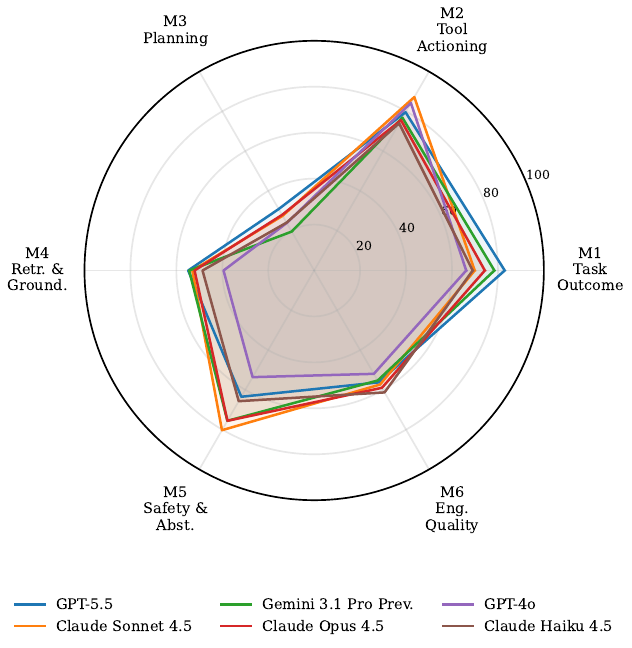}
\caption{\MOne{}--\MSix{} profiles of the top six models by \ERS{}. No
single model dominates every axis: \texttt{gpt-5.5} leads on task
outcome (\MOne{}), planning (\MThree{}), and retrieval grounding
(\MFour{}); Claude Sonnet~4.5 on tool actioning (\MTwo{}) and safety
and abstention (\MFive{}); and Claude Haiku~4.5 on engineering quality
(\MSix{}).}
\label{fig:m16-radar}
\end{minipage}
\end{figure*}

Figure~\ref{fig:m16-radar} overlays the top six models on
\MOne{}--\MSix{} as a radar chart, which makes the capability-level
specializations visible at a glance. Three observations follow.

First, top-line \ERS{} hides meaningful capability-level
specialization. \texttt{gpt-5.5} wins overall but is only third on
\MTwo{} (tool actioning, $79.56$) and fifth on \MFive{} (safety \&
abstention, $63.47$); Claude Sonnet~4.5, second overall by just
$0.36$ \ERS{} points, has the strongest tool-actioning and safety
profile ($87.17$ on \MTwo{} and $80.28$ on \MFive{}, the only safety
score above $76$). The strongest specialization signal sits below the
top tier: Claude Haiku~4.5's \MSix{} (Engineering Quality, $61.28$)
is the column-best in the entire roster, edging every frontier model
from rank~6. This is precisely the kind of decision the scalar \ERS{}
cannot surface on its own and that the \MOne{}--\MSix{} decomposition
was designed to expose.

Second, two axes have consistent frontier limits across all fourteen
models. \MThree{} (planning/workflow orchestration) tops out at
$30.89$ (\texttt{gpt-5.5}) and is below $25$ for nine of fourteen
models, driven almost entirely by \bench{UserBench}'s low
micro-average reward (no model exceeds $29$ on its travel envs; most
cluster in the $10$--$22$ range). The low absolute numbers are not a
scoring artifact: \bench{UserBench}'s \texttt{micro\_avg} averages
the per-turn reward across \emph{every} turn of an episode rather
than scoring only the final answer, so a model that wanders through
ten turns of generic chat before eventually recommending the correct
itinerary scores substantially below one that asks a single
clarifying question and resolves the task in three turns -- even
though both end with the correct answer. This is precisely the
step-efficiency signal \MThree{} is designed to measure, so the low
scores reflect a genuine finding rather than a quirk of how the
metric is constructed: no frontier model today is consistently
turn-efficient in multi-turn user-alignment tasks. Long-horizon
user-preference planning therefore remains an open frontier; current
LLMs do not yet saturate the multi-turn travel scenario. By contrast \MSix{} (engineering
quality) is compressed into a $51$--$61$ band, because its two routed
sub-metrics (\bench{LoCoBench}'s \texttt{software\_engineering} and
\texttt{code\_quality}) are saturated LLM-judged rubrics that reward
coherent output over strict correctness.

Third, the cost/capability boundary is fuzzier than parameter count
alone suggests. Gemini~2.5 Pro reaches \ERS{} $59.10$ -- virtually
tied with GPT-4o ($59.38$) and Claude Haiku~4.5 ($59.13$) -- powered
by an \MOne{} of $76.05$ that beats every other cost-tier model and
lands within $7$ points of \texttt{gpt-5.5}. At the lower frontier
edge, GPT-5 mini and Gemini~2.5 Flash are a near-tie at \ERS{} $51.91$
vs $51.36$, with GPT-5 mini's stronger \MFive{} ($54.17$) and \MSix{}
($56.12$) offsetting Gemini~2.5 Flash's higher \MThree{} ($24.72$).
At the smallest tier, GPT-5 nano holds the lowest-in-roster \MThree{}
($15.68$) and a near-floor \MOne{} ($42.38$) that leave it
second-from-last on overall \ERS{} despite a competitive \MTwo{}
($72.05$). The \MOne{}--\MSix{} decomposition therefore exposes
\emph{which} capability lets a smaller model close the gap, rather
than only \emph{whether} it does.

\section{Limitations}
\label{sec:limitations}

We note four limitations of this release. First, quick mode subsets
\bench{CRMArena}, \bench{HERB}, \bench{UserBench}, \bench{MCP-Universe},
\bench{MCPEval}, \bench{Benchforce}, and \bench{LoCoBench} to fit inside
the 180-minute budget, so sub-metric averages are computed over fewer
items and have wider confidence intervals. Readers should therefore
treat quick-mode \ERS{} as a
directional indicator and rely on deep-mode runs for release decisions.
Second, the $28/22/16/14/14/6$ weighting over \MOne--\MSix{} is
based on expert judgment and has not yet been
empirically recalibrated on \ebench{}; a calibration study that fits
the weights to observed enterprise risk (e.g., deployment incident
rates correlated with sub-metric patterns) is future work. Third,
\ERS{} is a scalar, and any single scalar over diverse capabilities
compresses information. Fourth,
parts of \bench{MCP-Universe}'s evaluator use LLM-based grading via the
configurable \texttt{EVALUATOR\_MODEL}, and judge choice and judge
drift can shift absolute numbers across runs even when the evaluated
LLM is identical. We fix the judge model in configuration so that
comparisons within a release are stable, but cross-release comparisons
should note the judge version.

\section{Conclusion}
\label{sec:conclusion}

\ebench{} unifies eight enterprise-facing LLM agent benchmarks behind a single
CLI and introduces a sub-metric-level routing framework (\MOne--\MSix{})
that produces an interpretable \ERS{} score without
collapsing capability-specific signal. The unit of evaluation is the
LLM: each contributing benchmark exercises the model through its own
native scaffold (ReAct, MCP client, multi-turn dialog, function-calling
harness). Runtime-budgeted quick and deep modes make the suite
tractable both for smoke tests and for release evaluations. We release the suite, configuration, and full quick-mode
results across fourteen frontier and cost-efficient LLMs spanning
the GPT, Claude, and Gemini families (Table~\ref{tab:main-results}).
The results expose capability-level specialization that is invisible
in any single benchmark's overall score: \texttt{gpt-5.5} leads
\ERS{} and sweeps \MOne{}, \MThree{}, and \MFour{}, but
Claude Sonnet~4.5 holds the column-best on \MTwo{} and \MFive{}, and
Claude Haiku~4.5 -- ranked sixth on \ERS{} -- holds the column-best
on \MSix{}. The source is available at
\url{https://github.com/SalesforceAIResearch/egi-bench}; we invite
extensions to the benchmark roster and contributions to the routing
table.

\bibliography{references}
\bibliographystyle{plainnat}

\appendix
\section{CLI Reference}
\label{sec:app-cli}

The \ebenchcli{} CLI is the preferred entry point. The lower-level
\texttt{run\_all.py} script is also supported; the CLI delegates to it
internally.

\paragraph{Installation.} \texttt{scripts/install\_all.sh} provisions the
root \ebenchcli{} CLI, the per-benchmark virtual environments, the LiteLLM
proxy venv, and the benchmark data in one idempotent step:
\begin{verbatim}
git clone https://github.com/SalesforceAIResearch/egi-bench.git
cd egi-bench
bash scripts/install_all.sh
source .venv/bin/activate
\end{verbatim}

\paragraph{Credentials.} A single \texttt{.env} file at the repository
root supplies all credentials. \texttt{OPENAI\_API\_KEY} is required even
when the model under test is non-OpenAI, because the judge and simulator
calls are pinned to OpenAI for apple-to-apple scoring; add the other
provider keys for whichever upstreams you route through:
\begin{verbatim}
OPENAI_API_KEY=your-openai-api-key   # also powers judge/simulator pins
ANTHROPIC_API_KEY=...                # optional, for Claude routes
GEMINI_API_KEY=...                   # optional, for Gemini routes
# optional deep-mode keys:
SERP_API_KEY=...           # MCP-Universe web_search
GOOGLE_MAPS_API_KEY=...    # MCP-Universe location_navigation
NPS_API_KEY=...            # MCPEval national_parks
\end{verbatim}

\paragraph{Model routing.} All LLM inference routes through a local
LiteLLM proxy on \texttt{http://localhost:8008/v1} (OpenAI-compatible).
Each \texttt{--model} is mapped to a provider via
\texttt{litellm\_proxy/litellm\_config.yaml}, so the same
\texttt{egi-bench run --model X} works for OpenAI, Anthropic, Gemini,
Bedrock, Vertex, Azure, OpenRouter, and self-hosted backends (vLLM,
Ollama, TGI). \texttt{egi-bench run} auto-spawns the proxy on demand;
power users can manage a long-lived proxy directly:
\begin{verbatim}
egi-bench proxy start     # long-lived proxy across runs
egi-bench proxy status
egi-bench proxy logs -f   # tail upstream provider errors
egi-bench proxy stop
\end{verbatim}

\paragraph{Benchmark environments and data.}
\begin{verbatim}
egi-bench doctor
egi-bench setup           # install per-benchmark .venv's + data
egi-bench setup --skip-data
egi-bench data            # refresh data only
\end{verbatim}

\paragraph{Running evaluations.}
\begin{verbatim}
egi-bench run --model gpt-4o-mini --mode quick --parallel
egi-bench run --model gpt-4o-mini --mode deep  --parallel --max-workers 2
egi-bench run --model gpt-4o-mini --mode quick \
              --benchmarks crmarena herb userbench
\end{verbatim}

\paragraph{Reporting.}
\begin{verbatim}
egi-bench report                                # latest aggregate
egi-bench report --rescore --model <m> --mode quick
egi-bench report --rescore --model <m> --mode quick \
        --egi-weights M1=0.20,M5=0.30
egi-bench clean
\end{verbatim}

\texttt{report --rescore} re-derives every benchmark's sub-metric values
from cached raw outputs and re-runs the aggregator, enabling routing or
weight changes without re-running inference.

\section{Benchmark Scope and Metric Coverage Tables}
\label{sec:app-bench-tables}

This appendix collects the two reference tables summarizing the eight
benchmarks: Table~\ref{tab:bench-overview} lists each benchmark's
domain, task counts under quick and deep modes, and its native primary
metric; Table~\ref{tab:coverage} shows which benchmarks (and which
sub-metrics) contribute to each of the \MOne--\MSix{} enterprise
metrics.

\begin{table*}[t]
\centering
\small
\setlength{\tabcolsep}{5pt}
\renewcommand{\arraystretch}{1.05}
\caption{The eight benchmarks unified under \ebench{}. Task counts reflect the
quick and deep mode profiles defined in \texttt{benchmark\_config.yaml}. Full
benchmark descriptions are in Section~\ref{sec:framework}.}
\label{tab:bench-overview}
\begin{tabular}{@{}
  >{\raggedright\arraybackslash}p{2.4cm}
  >{\raggedright\arraybackslash}p{3.2cm}
  >{\raggedright\arraybackslash}p{5.6cm}
  >{\raggedright\arraybackslash}p{4.4cm}
@{}}
\toprule
Benchmark & Domain & Tasks (quick / deep) & Primary metric \\
\midrule
\bench{CRMArena}     & CRM operations          & 9 types $\times$ 6 / $\times$ 3            & Task accuracy \\
\bench{HERB}         & Enterprise search       & 80 Qs / full                                & F1 $+$ content correctness \\
\bench{PersonaBench} & Personal data QA        & 3 people $\times$ 2 / $\times$ 4 noises     & QA accuracy \\
\bench{UserBench}    & Travel planning         & 3 envs $\times$ 60 / all envs               & Micro-avg.\ reward \\
\bench{MCP-Universe} & MCP tool use            & 1 dom.\ $\times$ 20 / 3 dom.                & Sub-eval pass rate \\
\bench{MCPEval}      & MCP tool use            & 1 dom.\ $\times$ 50 / 3 dom.\ $\times$ 50   & Flex.\ avg.\ score \\
\bench{Benchforce}   & Function calling        & 1 env $\times$ 40 / 2 envs full             & Accuracy \\
\bench{LoCoBench}    & Long-context SWE        & 30 easy / 500 full                          & 0--5 rubric \\
\bottomrule
\end{tabular}
\end{table*}

\begin{table*}[t]
\centering
\small
\setlength{\tabcolsep}{6pt}
\renewcommand{\arraystretch}{1.15}
\caption{Which benchmarks contribute to each enterprise metric after
fine-grained decomposition. Every benchmark feeds at least one bucket and
every bucket receives at least two routed sub-metrics; \MSix{} is the
only bucket whose sub-metrics come from a single benchmark
(\bench{LoCoBench}).}
\label{tab:coverage}
\begin{tabular}{@{}l >{\raggedright\arraybackslash}p{15.2cm}@{}}
\toprule
Metric & Contributing benchmarks (sub-metric sources) \\
\midrule
\MOne{}  & \bench{CRMArena} (NCR, HTU, TCU, TII, BRI, MTA), \bench{Benchforce} (\texttt{accuracy\_all}), \bench{LoCoBench} (functional\_correctness) \\
\MTwo{}  & \bench{MCP-Universe} (per-domain pass rate), \bench{MCPEval} (flexible average score), \bench{Benchforce} (finished ratio) \\
\MThree{}& \bench{UserBench} (micro\_avg), \bench{LoCoBench} (longcontext\_utilization) \\
\MFour{} & \bench{CRMArena} (NED, KQA), \bench{HERB} (per-entity F1, content correctness), \bench{PersonaBench} (QA accuracy) \\
\MFive{} & \bench{CRMArena} (PVI), \bench{HERB} (unanswerable null rate), \bench{Benchforce} (unsafe-failure gap) \\
\MSix{}  & \bench{LoCoBench} (software\_engineering, code\_quality) \\
\bottomrule
\end{tabular}
\end{table*}

\section{Fine-Grained Sub-Metric Routing}
\label{sec:routing}

The routing table (Table~\ref{tab:routing-matrix}) is the operational
heart of \ebench{}. Every row lists a sub-metric that a benchmark natively
reports and the enterprise bucket (\MOne--\MSix{}) it feeds. No
benchmark's overall score is consumed directly.

\paragraph{Why sub-metric routing.} Different sub-metrics inside the same
benchmark often probe very different capabilities. In \bench{CRMArena},
PVI (Policy Violation Identification) is a direct safety signal, KQA is
a retrieval-grounded knowledge task, and NCR is a business-action
routing task; averaging the three loses the distinction that
enterprise stakeholders care about. By routing PVI into \MFive{}, KQA
and NED into \MFour{}, and the remaining six task types into \MOne{},
we avoid dilution across capabilities and retain diagnostic signal at
the bucket level. The same logic applies to \bench{HERB}, whose
grounded per-entity F1 and content correctness feed \MFour{} while its
unanswerable null rate feeds \MFive{} as a dedicated anti-hallucination
signal.

\begingroup
\footnotesize
\setlength{\tabcolsep}{5pt}
\renewcommand{\arraystretch}{1.10}
\begin{longtable}{@{}
  >{\raggedright\arraybackslash}p{3.0cm}
  >{\raggedright\arraybackslash}p{5.4cm}
  >{\centering\arraybackslash}p{1.0cm}
  >{\raggedright\arraybackslash}p{6.5cm}
@{}}
\caption{Fine-grained sub-metric routing. Each row shows a native sub-metric
reported by a benchmark, the enterprise metric (\MOne--\MSix) it is routed
into, and a short rationale. No benchmark's overall score is consumed
directly; the routing is the unit of aggregation.}
\label{tab:routing-matrix} \\
\toprule
Benchmark & Sub-metric & Bucket & Rationale \\
\midrule
\endfirsthead
\multicolumn{4}{l}{\emph{Table~\ref{tab:routing-matrix} continued from previous page}} \\
\toprule
Benchmark & Sub-metric & Bucket & Rationale \\
\midrule
\endhead
\midrule \multicolumn{4}{r}{\emph{continued on next page}} \\
\endfoot
\bottomrule
\endlastfoot

\bench{CRMArena}     & NCR (Case Routing)                      & \MOne   & Business action correctness \\
                     & HTU (Handle Time)                       & \MOne   & Analytical task outcome \\
                     & TCU (Transfer Count)                    & \MOne   & Analytical task outcome \\
                     & TII (Top Issue ID)                      & \MOne   & Analytical task outcome \\
                     & BRI (Best Region ID)                    & \MOne   & Analytical task outcome \\
                     & MTA (Monthly Trend)                     & \MOne   & Task outcome \\
                     & NED (Named Entity Disamb.)              & \MFour  & Entity resolution is grounded retrieval \\
                     & KQA (Knowledge QA)                      & \MFour  & Knowledge retrieval correctness \\
                     & PVI (Policy Violation ID)               & \MFive  & Direct policy / safety signal \\
\addlinespace
\bench{HERB}         & Per-entity F1 (person/url/pr/company)   & \MFour  & Extraction / grounding quality \\
                     & Content correctness                     & \MFour  & Evidence-grounded answering \\
                     & Unanswerable null rate                  & \MFive  & Primary anti-hallucination signal \\
\addlinespace
\bench{PersonaBench} & Per-(noise\,$\times$\,type) QA accuracy  & \MFour  & Context-grounded personalized retrieval \\
\addlinespace
\bench{MCP-Universe} & Pass rate: \texttt{financial\_analysis} & \MTwo   & Tool actioning in domain (quick) \\
                     & Pass rate: \texttt{web\_search}         & \MTwo   & Tool actioning in domain (deep) \\
                     & Pass rate: \texttt{location\_nav.}      & \MTwo   & Tool actioning in domain (deep) \\
\addlinespace
\bench{MCPEval}      & Flexible per-domain average score        & \MTwo   & Tool actioning reliability \\
\addlinespace
\bench{UserBench}    & Per-env \texttt{micro\_avg}             & \MThree & Multi-turn planning quality \\
\addlinespace
\bench{Benchforce}   & \texttt{accuracy\_all} per env          & \MOne   & Task outcome correctness \\
                     & \texttt{finished\_entries/total}        & \MTwo   & Operational actioning reliability \\
                     & \texttt{acc\_finished}$-$\texttt{acc\_all} gap & \MFive  & Unsafe-failure proxy \\
\addlinespace
\bench{LoCoBench}    & functional\_correctness                  & \MOne   & Solves the task \\
                     & software\_engineering                    & \MSix   & Engineering discipline \\
                     & code\_quality                            & \MSix   & Maintainability \\
                     & longcontext\_utilization                 & \MThree & Long-horizon planning over context \\
\end{longtable}
\endgroup

\paragraph{Normalization and aggregation.} Sub-metrics are first rescaled
to $[0, 100]$ with each benchmark's native conventions (F1 $\in [0,1]$
becomes $[0, 100]$; pass rates stay proportional). Lower-is-better
signals are converted to higher-is-better before routing (e.g.\
\bench{Benchforce}'s reliability gap enters \MFive{} as
$100 - (\texttt{acc\_finished} - \texttt{acc\_all})$). Within each bucket we default to
an equal-weighted mean of all routed sub-metrics; \texttt{benchmark\_config.yaml}
supports per-sub-metric weights when a principled override is warranted.

\paragraph{Missing-data handling.} If a sub-metric is unavailable (e.g.\
a benchmark was skipped), we reweight \emph{within} its bucket only,
never globally. If a whole benchmark run fails, all of its contributed
sub-metrics are dropped from their buckets and every affected bucket is
renormalized locally. The per-run aggregate report explicitly flags which
sub-metrics were missing so readers can judge coverage.

\paragraph{Routing is auditable.} The routing is stored declaratively in
\texttt{benchmark\_config.yaml} so reviewers can override it and rerun
aggregation without touching code. Section~\ref{sec:modes} describes the
CLI's \texttt{report --rescore} subcommand that re-derives \ERS{} from
stored raw outputs when the routing changes.

\section{Extended Routing Rationale}
\label{sec:app-routing}

This appendix expands on the routing table in
Section~\ref{sec:routing} with the reasoning behind the more debatable
assignments.

\paragraph{\bench{CRMArena}.} Six analytical task types (NCR, HTU, TCU,
TII, BRI, MTA) go to \MOne{} because each produces a single
business-facing answer whose correctness is measured by exact or
structured match. NED (Named Entity Disambiguation) goes to \MFour{}
because the ground truth depends on resolving entities from retrieved
records; KQA (Knowledge QA) similarly tests retrieval-grounded QA. PVI
(Policy Violation Identification) goes to \MFive{} because it is the
only task type whose ground truth is a policy judgment --- a direct
safety signal.

\paragraph{\bench{HERB}.} Per-entity F1 (over person, url, pr, and
company fields) and content correctness measure grounded retrieval
quality and feed \MFour{}. The unanswerable null rate measures whether
the LLM abstains when no answer exists; it feeds \MFive{} and is in
practice our primary anti-hallucination signal.

\paragraph{\bench{MCPEval}.} The routed signal is the flexible
per-domain average score, which feeds \MTwo{} (tool actioning
reliability). The scorer also computes decomposed components ---
tool-name match, parameter match, order, and strict success --- but
these are not yet surfaced to the aggregate runner. When they are, the
order component is slated to route to \MThree{} (tool sequencing is
planning) and the strict success rate to \MFive{} (all-or-nothing, hence
sensitive to drift and over-action).

\paragraph{\bench{LoCoBench}.} \texttt{functional\_correctness} goes to
\MOne{} (solves the task). \texttt{software\_engineering} and
\texttt{code\_quality} feed \MSix{} (engineering discipline;
maintainability). \texttt{longcontext\_utilization} feeds \MThree{}
because the dimension measures long-horizon reasoning and planning over
extended context, which is a planning/orchestration signal rather than a
retrieval one.

\paragraph{Debatable cases.} The routing of
\texttt{longcontext\_utilization} to \MThree{} rather than \MFour{} is
the most debatable assignment among the currently active routes; the
planned \bench{MCPEval} component routing (order $\to$ \MThree{}, strict
$\to$ \MFive{} rather than \MTwo{}) is a second. All routings are
recorded in \texttt{benchmark\_config.yaml} and are overridable via
\texttt{egi-bench report --rescore} without re-running inference.

\section{Execution Modes}
\label{sec:modes}

\ebench{} exposes two execution modes, selected per run via
\texttt{--mode quick|deep}. Modes are defined by \emph{total wall-clock},
not by coverage heuristics:

\begin{itemize}
  \item \textbf{Quick mode} targets a total wall-clock under ${\sim}180$~min
  with \texttt{--parallel --max-workers 2}.
  \item \textbf{Deep mode} targets ${\sim}12$~h of wall-clock with similar
  parallelism.
\end{itemize}

Coverage decisions (which categories, how many tasks per category) fall
out of a per-benchmark runtime budget rather than being picked by hand.
``Fast enough'' benchmarks keep their original upstream configuration in
both modes; slower benchmarks get subsetted. Concrete wall-clock numbers
depend on the host machine and the inference endpoint, so we define
modes by budget targets rather than by a fixed per-benchmark runtime
table.

\paragraph{Subsetting rules.} When a benchmark must be subsetted:

\begin{enumerate}
  \item Keep at least one task per existing category so every reported
  sub-metric retains signal.
  \item Reduce \emph{tasks per category} before dropping a category,
  since dropping a category also drops a bucket contributor.
  \item If a category must be dropped in quick mode, prefer the one
  whose routed bucket is most redundantly covered elsewhere (e.g., a
  \bench{MCP-Universe} domain whose \MTwo{} signal is already supplied
  by \bench{MCPEval}/\bench{Benchforce}).
  \item Record subset definitions in \texttt{benchmark\_config.yaml}; do
  not hard-code them in \texttt{run\_all.py}.
\end{enumerate}

\paragraph{CLI surface.} The preferred entry points are:

\begin{verbatim}
egi-bench doctor        # dependency check
egi-bench setup         # install per-benchmark venvs + data
egi-bench run --model <m> --mode {quick,deep} --parallel
egi-bench report                                  # latest
egi-bench report --rescore --model <m> --mode <m> # re-aggregate
\end{verbatim}

\texttt{report --rescore} re-derives every benchmark's sub-metric values
from cached raw outputs and re-runs the \MOne--\MSix{} aggregator. This
means routing or weighting changes can be evaluated without re-running
inference, which is important for framework stability as
Table~\ref{tab:routing-matrix} evolves.

\paragraph{Credentials.} Both modes require \texttt{OPENAI\_API\_KEY}
(it powers the OpenAI routes and the pinned judge/simulator calls, so it
is needed even when the model under test is non-OpenAI), plus whichever
provider keys correspond to the upstreams being routed through the
LiteLLM proxy (e.g.\ \texttt{ANTHROPIC\_API\_KEY},
\texttt{GEMINI\_API\_KEY}). Deep mode additionally benefits from
\texttt{SERP\_API\_KEY} and \texttt{GOOGLE\_MAPS\_API\_KEY}
(\bench{MCP-Universe}) and \texttt{NPS\_API\_KEY} (\bench{MCPEval}).
Missing optional keys cause the affected domain to be skipped with a
note, not a failure; the resulting aggregate reports reduced coverage.

\section{Sub-Metric Catalog}
\label{sec:app-catalog}

The catalog below lists all sub-metrics currently emitted by each
benchmark's scorer in \ebench{}. Sub-metrics marked ``not routed'' are
filtered out of \MOne--\MSix{} aggregation as internal diagnostics.

\subsection*{\bench{CRMArena}}
Sub-metrics: NCR, HTU, TCU, TII, BRI, MTA, NED, KQA, PVI. All routed
(six into \MOne{}, two into \MFour{}, one into \MFive{}).

\subsection*{\bench{HERB}}
Sub-metrics: per-entity F1 across \{person, url, pr, company\}
(\MFour{}), content correctness (\MFour{}), unanswerable null rate
(\MFive{}).

\subsection*{\bench{PersonaBench}}
Sub-metrics by noise level ($0.0, 0.3, 0.5, 0.7$) and question type
(Basic information, Preference, Subjective, Social). The per-(noise
$\times$ question-type) QA accuracies all feed \MFour{}; the individual
\texttt{result\_noise\_x\_y} keys are retained for diagnosis but not
routed separately.

\subsection*{\bench{UserBench}}
Sub-metrics: \texttt{micro\_avg} per env (\texttt{travel22},
\texttt{travel33}, \texttt{travel44}) $\to$ \MThree{}. Filtered (not
routed): \texttt{micro\_max}, \texttt{avg\_number\_of\_1},
\texttt{avg\_number\_of\_08}.

\subsection*{\bench{MCP-Universe}}
Sub-metrics: per-domain sub-evaluator pass rate (\MTwo{}). Quick mode
exercises \texttt{financial\_analysis}; deep mode adds
\texttt{web\_search} and \texttt{location\_navigation}, gated on API-key
availability.

\subsection*{\bench{MCPEval}}
Routed sub-metric: the flexible per-domain \texttt{average\_score}
(\MTwo{}). The scorer also emits decomposed components (flexible
tool-name, parameter-match, order, and strict success rate) into each
domain's \texttt{*\_analysis.json}, but these are not yet surfaced to the
aggregate runner; when they are, \texttt{order} is slated for \MThree{}
and \texttt{strict.success\_rate} for \MFive{}.

\subsection*{\bench{Benchforce}}
Sub-metrics per env (hospital, hr\_software): \texttt{accuracy\_all}
(\MOne{}), \texttt{finished\_entries}/\texttt{total} (\MTwo{}),
\texttt{accuracy\_finished}$-$\texttt{accuracy\_all} gap (\MFive{}).
Filtered (not routed): token and turn efficiency --- orthogonal to
correctness.

\subsection*{\bench{LoCoBench}}
Sub-metrics routed: \texttt{functional\_correctness} (\MOne{}),
\texttt{software\_engineering} (\MSix{}), \texttt{code\_quality}
(\MSix{}), \texttt{longcontext\_utilization} (\MThree{}). Filtered: the
17 finer-grained dimensions and generation statistics used internally
by LoCoBench's rubric.

\section{Reproducibility}
\label{sec:app-repro}

\paragraph{Software versions.} Python~3.12+ and Node.js~20+;
dependencies are installed per-benchmark via \texttt{uv venv}/\texttt{uv
pip install}, and each benchmark's exact lockfile lives in its
subdirectory. The whole stack --- root CLI, per-benchmark venvs, the
LiteLLM proxy venv, and benchmark data --- is provisioned from the
repository root via the idempotent \texttt{bash scripts/install\_all.sh}.

\paragraph{Proxy and models.} All LLM inference routes through a local
LiteLLM proxy on \texttt{http://localhost:8008/v1} (OpenAI-compatible);
the model-to-provider mapping is declared in
\texttt{litellm\_proxy/litellm\_config.yaml}. To keep scores comparable
across models, that config \emph{pins} the judge/simulator model names
(e.g.\ \texttt{gpt-4o}, \texttt{gpt-4.1}) to OpenAI regardless of the
model under test, so \texttt{OPENAI\_API\_KEY} must be set even for
non-OpenAI runs. The evaluator model for \bench{MCP-Universe} is
configured via \texttt{EVALUATOR\_MODEL} in \texttt{.env} (defaults to
\texttt{gpt-4.1}).

\paragraph{Seeds.} \texttt{benchmark\_config.yaml} sets
\texttt{sample\_seed: 42} for subsetting where applicable
(\bench{CRMArena}, \bench{MCP-Universe}). Other benchmarks use
deterministic ordering over their native task files.

\paragraph{Quick-mode subset definitions.} For each benchmark, the quick
profile is recorded as follows (excerpted from
\texttt{benchmark\_config.yaml}):

\begin{itemize}
  \item \bench{CRMArena}: all 9 task types, up to 6 tasks per category.
  \item \bench{HERB}: answerable and unanswerable modes; 80 questions total.
  \item \bench{PersonaBench}: 3 people, communities \{0, 1\}, noise
        levels \{0.0, 0.7\}, \texttt{all-MiniLM-L6-v2} retriever.
  \item \bench{UserBench}: envs \{travel22, travel33, travel44\}, up to
        60 tasks/env, 20 turns, pass\_k 1, temperature 1.0.
  \item \bench{MCP-Universe}: domain \{financial\_analysis\}, 20 tasks.
  \item \bench{MCPEval}: domain \{airbnb\}, 50 tasks, 20 turns.
  \item \bench{Benchforce}: env \{hospital\}, 40 entries, 15 turns.
  \item \bench{LoCoBench}: difficulty \texttt{easy}, 30 scenarios,
        1800\,s timeout.
\end{itemize}

Deep-mode counterparts expand each list to its upstream default.

\paragraph{Artifacts.} Each run produces
\texttt{results\_aggregated/<model>\_<mode>\_<timestamp>.json} plus
per-benchmark logs under \texttt{results\_aggregated/logs/}. The JSON
contains the full four-level metrics structure: scalar \ERS{},
\MOne--\MSix{} with per-contributor breakdown, per-benchmark sub-metric
values, and routing metadata.

\paragraph{Reproducing a quick-mode run.} The end-to-end command
sequence is identical across the fourteen-model roster; only the
\texttt{--model} string changes. Claude Sonnet~4.5 is shown as an
example.

\begin{verbatim}
# .env holds OPENAI_API_KEY (judge/simulator pins) plus the provider
# key for the model under test, e.g. ANTHROPIC_API_KEY.
# egi-bench run auto-spawns the LiteLLM proxy on localhost:8008.
egi-bench run \
    --model claude-sonnet-4-5 \
    --mode quick --parallel --max-workers 2
egi-bench report
\end{verbatim}

\end{document}